\documentclass[fleqn,10pt]{wlscirep}
\usepackage[utf8]{inputenc}
\usepackage[T1]{fontenc}
\usepackage{amsmath}
\usepackage{amssymb}
\usepackage{graphicx}
\usepackage{booktabs}
\usepackage{hyperref}

\newcommand{\U}{\bigtriangleup}
\newcommand{\D}{\bigtriangledown}
\newcommand{\E}{\square}
\newcommand{\A}{A}
\newcommand{\K}{\mathsf{K}}

\title{Kendall transformation: a robust representation of continuous data for information theory}
\author[1,*]{Miron Bartosz Kursa}

\affil[1]{Interdisciplinary Centre for Mathematical and Computational Modelling\\ University of Warsaw}
\affil[*]{M.Kursa@icm.edu.pl}

\begin{abstract}
Kendall transformation is a conversion of an ordered feature into a vector of pairwise order relations between individual values.
This way, it preserves ranking of observations and represents it in a categorical form.

Such transformation allows for generalisation of methods requiring strictly categorical input, especially in the limit of small number of observations, when discretisation becomes problematic.
In particular, many approaches of information theory can be directly applied to Kendall-transformed continuous data without relying on differential entropy or any additional parameters.
Moreover, by filtering information to this contained in ranking, Kendall transformation leads to a better robustness at a reasonable cost of dropping sophisticated interactions which are anyhow unlikely to be correctly estimated.

In bivariate analysis, Kendall transformation can be related to popular non-parametric methods, showing the soundness of the approach.
The paper also demonstrates its efficiency in multivariate problems, as well as provides an example analysis of a real-world data.
\end{abstract}


\begin{document}

\flushbottom
\maketitle
\thispagestyle{empty}

\section{Introduction}
Information theory~\cite{Shannon1948} is a powerful framework utilised in many branches of statistics and machine learning.
It is used, among others, for association testing~\cite{Smith2015}, feature selection~\cite{Brown2012}, network reconstruction~\cite{Margolin2006} and clustering~\cite{Brown1992}.
The known deficiency of the theory, though, is that it is well defined for discrete probability distributions, yet there is no unequivocally proper generalisation over continuous distributions which would retain important properties without causing substantial theoretical or practical issues.

A common approach here is to simply discretise data before information-theoretic analysis, and treat it as discrete afterwards; this process is lossy and can be done in numerous ways, however, bringing additional burden of heuristics and hyper-parameters which can critically influence the conclusion, especially in the small data limit.
On the other hand, the natural theoretical generalisation is differential entropy; unfortunately, while very useful on its own, it violates many useful properties of discrete Shannon entropy and related quantities, seriously impacting applicability of more sophisticated methods developed for discrete cases.
Moreover, on a practical side, this approach requires estimation of continuous distributions underlying analysed data, which may easily become a highly non-trivial task relying on cumbersome heuristics, indifferent from the discretisation way.
Henceforth, it is desirable to look for more robust and generic approaches.

In this paper, I will follow the idea behind Kendall correlation~\cite{Kendall1938}, namely to represent a continuous variable as a graph of greater-than relations between each pair of its values, and to measure association of variables in terms of their similarity in such representation.
Still, the said graph can be thought of not just as an intermediate artefact, but an useful concept on its own.
In particular, the list of its edges, when ordered in some arbitrary yet common way, can be written using at most three states, hence can be treated as a categorical encoding of the ranking of an original variable.

I will argue that such a variable is a faithful representation of its original in many information-based inquiries into the data.
Therefore such conversion, later called \textit{Kendall transformation}, is a reliable, parameter-less alternative to both discretisation and elaborate continuous entropy estimators.

\begin{figure}
 \includegraphics[width=\textwidth]{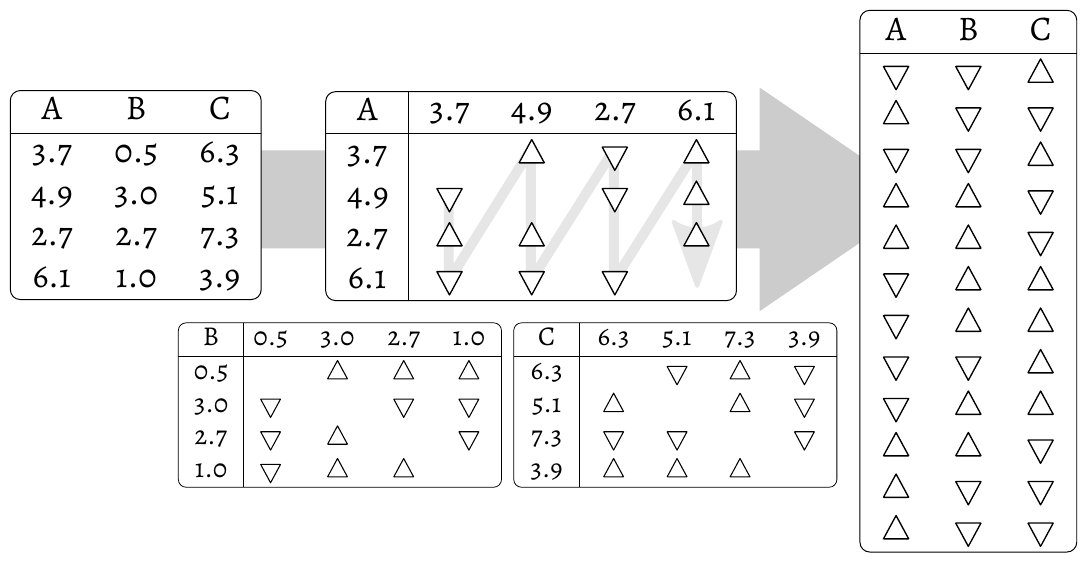}
 \caption{\label{fig:ex} Kendall transformation of a toy information system (left).
 Features A, B and C are converted into matrices of relations between values (centre).
 Matrices are flattened in an arbitrary but consistent manner and collected into a transformed system (right).}
\end{figure}

\section{Kendall transformation}
Let us assume some way of ordering all $m=n(n-1)$ possible pairs from a set $\{1..n\}$, with $(a_j,b_j \neq a_j)$ denoting $j$-th such pair. 
For an arbitrary $n$-element vector $x_i\in X^n$, where $X$ is a set with total order, Kendall transformation is defined as 
\begin{equation}
{\K(x_i)}_j:=
\begin{cases}
\U & \text{if } x_{a_j}<x_{b_j}\\
\D & \text{if } x_{a_j}>x_{b_j}\\
\E & \text{if } x_{a_j}=x_{b_j}\\
\end{cases}
\in {\{\U,\D,\E\}}^m. 
\end{equation}
By extension, Kendall transformation of an information system is an information system in which every attribute has been Kendall-transformed.
Figure~\ref{fig:ex} presents an illustrative example of transformation of a small, toy system.
It is trivial to note that $\K(x)=\K(f(x))$ for any strictly increasing $f$; henceforth, the proposed transformation is lossy, but faithfully preserves observation ranking.

\subsection{Bivariate analysis}
Let us further assume for a moment that there are no ties between values, hence that the equal state does not occur.
With two attributes $x_i$ and $y_i$,  we say that pair $j$ is concordant if $\K{(\vec{x})}_j=\K{(\vec{y})}_j$ and discordant otherwise.
The Kendall correlation coefficient $\tau$ is then the difference between the number of concordant and discordant pairs, normalised by $m$.\footnote{One should note that $\tau$ is quantised into $m/2+1$ states; in particular, $\tau=0$ is only possible for $n$ or $n-1$ divisible by 4.}
Furthermore, the entropy of a Kendall-transformed vector is $\log(2)$, and the mutual information (MI) between two transformed variables $I^\K$ is a function of a Kendall correlation $\tau$ between them, namely
\begin{equation}
\label{eq:kenmi}
I^\K(\tau)=\tau\log\sqrt{\frac{1+\tau}{1-\tau}}+\log\sqrt{1-\tau^2}.
\end{equation}
It is an even function, strictly increasing for $\tau>0$.
Moreover, it can be also considered as an extension of a simple, commonly used formula connecting MI and correlation coefficient $\rho$,
\begin{equation}
\label{eq:cormi}
I^G(\rho)=-\log \sqrt{1-\rho^2}.
\end{equation}
It is derived by subtracting differential entropy of a bivariate normal distribution from a sum of differential entropies of its marginals, with a constant factor $\log\sqrt{2\pi e}$ omitted, and, although valid only for the Pearson correlation, it is often used with other coefficients, usually Spearman's~\cite{DeJay2013}.
Both functions behave similarly for small absolute values of $\tau$ and $\rho$.
On the other hand, when respective correlation coefficient approaches 1 or -1, $I^\K$ achieves maximum, while $I^G$ diverges, causing problems in certain use cases, especially when highly correlated features are of interest.

As with Pearson or Spearman~\cite{Spearman1904} correlation, on Kendall-transformed information systems we can only detect monotonic relationships; if, say, $x\sim\mathcal{U}(-1,1)$, the relation between $x$ and $x^2$ will be lost.
I will argue that it is well justified constraint in small-$n$ problems, however.
Under the null hypothesis of interaction testing, independence between variables, the probability of each joint state is a product of its marginal probabilities.
It is very unlikely to get a small sample  of such symmetry, though, hence the agreement with null may easily become less likely than with the alternative hypothesis.
Obviously, we may apply corrections for this phenomenon, but for a cost of severely hindered sensitivity.
On the other hand, even among short sequences, the probability of randomly getting a sorted one is minimal ($1/n!$ without ties), hence the null of non-monotonicity is much more robust.

\begin{figure}
 \includegraphics[width=\textwidth]{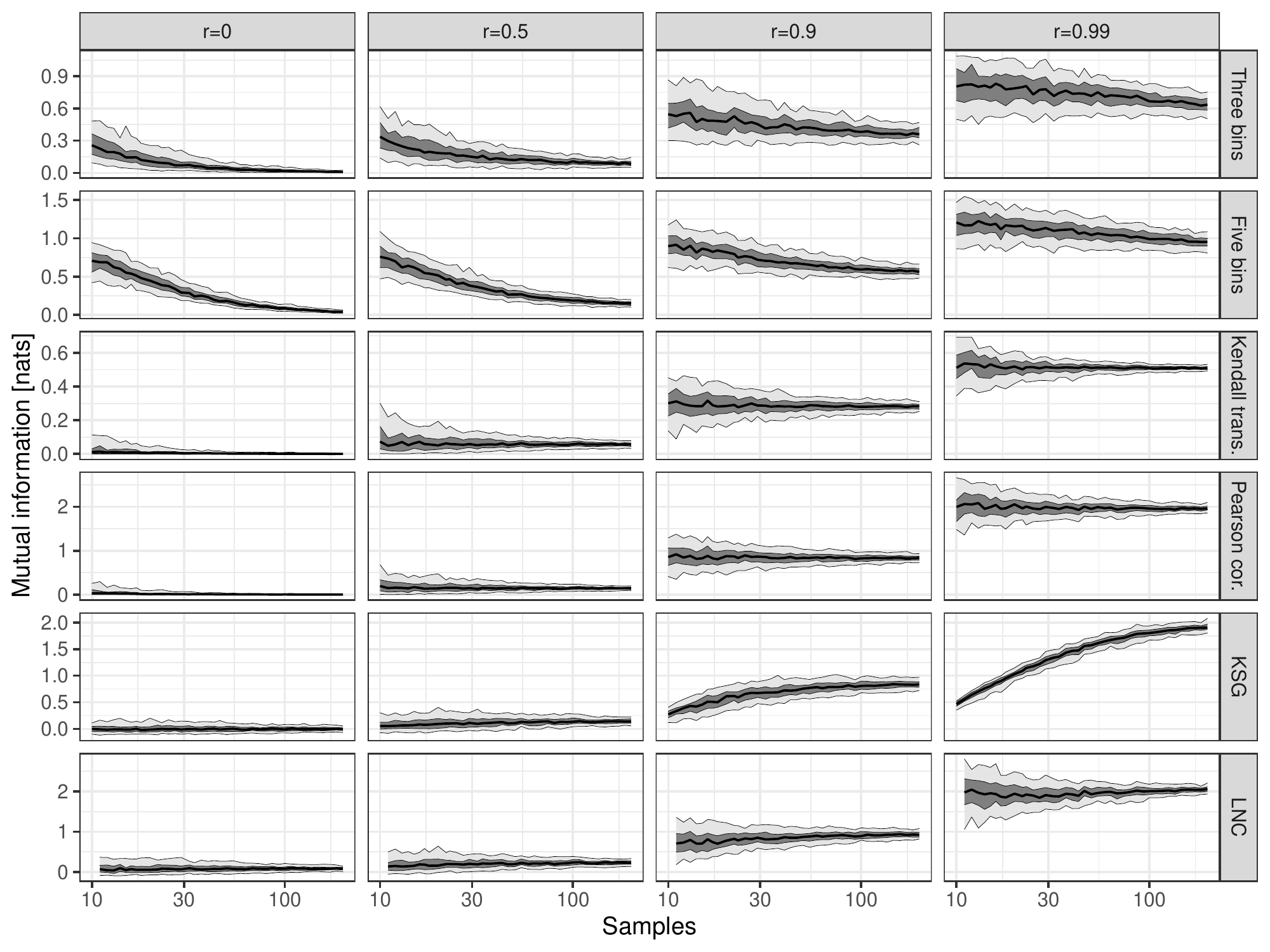}
 \caption{\label{fig:biv} Mutual information between two variables drawn from a bivariate normal distribution with correlation $r$, calculated with different methods. The lines show 5th, 25th, 50th, 75th and 95th percentile over 100 realisations.}
\end{figure}

Let us empirically investigate the properties of $I^\K$ on a synthetic data drawn from bivariate normal distribution; we set the marginal distributions to $\mathcal{N}(0,1)$, and covariance to $r$.
For a comparison, we will use five other methods of estimating MI: discretisation into three or five bins of equal width, Pearson correlation coefficient with Equation~\ref{eq:cormi}, finally two versions of k-NN estimator, variant 2 of the Kraskov estimator~\cite{Kraskov2004} with $k=5$ (KSG) and its extension corrected for local non-uniformity \cite{Gao2015} with $k=10$ and $\alpha=0.65$ (LNC).
When not noted otherwise, a maximal likelihood estimator of entropy is used. 
The results of this experiment, repeated for four values of $r$, various sample sizes $n$ and over 100 random replicates, are collected on Figure~\ref{fig:biv}.

We can see that discretisation approach leads to a very poor convergence, which makes such estimators unsuitable for low $n$ cases; this is especially pronounced for $r=0$, which in practice would lead to many false positives in association sweeps.
In contrast, $I^\K$ provides nearly unbiased results even for smallest values of $n$, as well as a relatively small variance.

The remaining methods estimate differential MI, so $-\log\sqrt{1-r^2}$.
The one based on Pearson correlation, which behaviour closely resembles this of $I^\K$, yet with a smaller variance.
Still, this is due to the fact that we have fully satisfied strong assumptions this estimator relies on; it won't be nearly as effective in a general case.
The KSG estimator has a very low variance, and provides a bell-shaped distribution for estimates of $r=0$, which is handy for independence testing.
Still, in highly correlated cases it exhibits its known deficiency of a very slow convergence.
The LNC estimator, on the other hand, converges fast for any $r$; it is visibly biased in the independence case, however, which is likely to hurt specificity.

Clearly, Kendall transformation is the most versatile and robust among investigated solutions; it works reliably over the entire range of analysed cases, takes no parameters and is never substantially inferior to the best method.

\subsection{Ties}
As mentioned earlier, the entropy of a Kendall-transformed variable is $\log(2)$ if there are no ties, regardless of the distribution of the original.
This becomes intuitive given that this transformation, similarly to ranking, wipes scale information and retains only order; hence, it effectively converts any input distribution into an uniform one.
Tied values cannot be separated by a such process, hence the resulting effective distribution in a general case is a mixture of an uniform distribution and Dirac deltas located over tied values, which is a richer, more complex structure.
In a similar fashion, the introduction of ties generate $\E$ states in the transformed variable which first increases its entropy, up to $\log(3)$, when the complexity contributions of uniform and discrete components become balanced.
Additional ties effectively convert the distribution into a discrete one, decreasing the entropy of the transformation.
Finally, due to coalescence of values, only one state remains both in the original variable and its Kendall transformation, and entropies of both become 0.

Naturally, the correspondence between actual entropy of a discrete variable and the entropy of a Kendall transformation of its numerical encoding holds only in a constant and binary case; otherwise the order in which states are encoded becomes important.
Henceforth, Kendall transformation is directly applicable to numeric, ordinal and binary features, and can provide a viable representation of ties when they are not numerous enough to dominate the ordinal nature of a feature.
The proper handling of arbitrary categorical data in such a framework is a subject for further research, though, by a simple extension, we may analyse such features by breaking them into a set of category-vs-other indicator features.

One should note, though, that the above reasoning applies to actual ties, i.e., pairs of values that are indistinguishable, like two days without rain; as opposed to two days with such a similar amount of precipitation that the resolution of the sensor is insufficient to differentiate them.
In the latter case it is better to break such artificial ties using random jitter or to treat comparisons between them within the Kendall-transformed variable as missing observations.

Interestingly, when $x$ is ordered and contains no ties and $y$ is binary, $I^\K$ also corresponds to a popular measure of association.
Namely, it is a function of $\A$ defined as the area under the receiver operating characteristic (ROC) curve,
\begin{equation}
\label{eq:aurocmi}
I^\K(\A;a;b)=\frac{2ab}{n(n-1)}\left(\A\log\frac{\A}{1-\A}+\log(2-2\A) \right),
\end{equation}
normalised by sizes of both classes, $a$ and $b=n-a$.
This way, Kendall transformation is also connected with the Mann-Whitney-Wilcoxon test~\cite{Mann1947}, as its statistic $U=ab(1-\A)$.

\subsection{Multivariate analysis}
\begin{figure}
 \includegraphics[width=\textwidth]{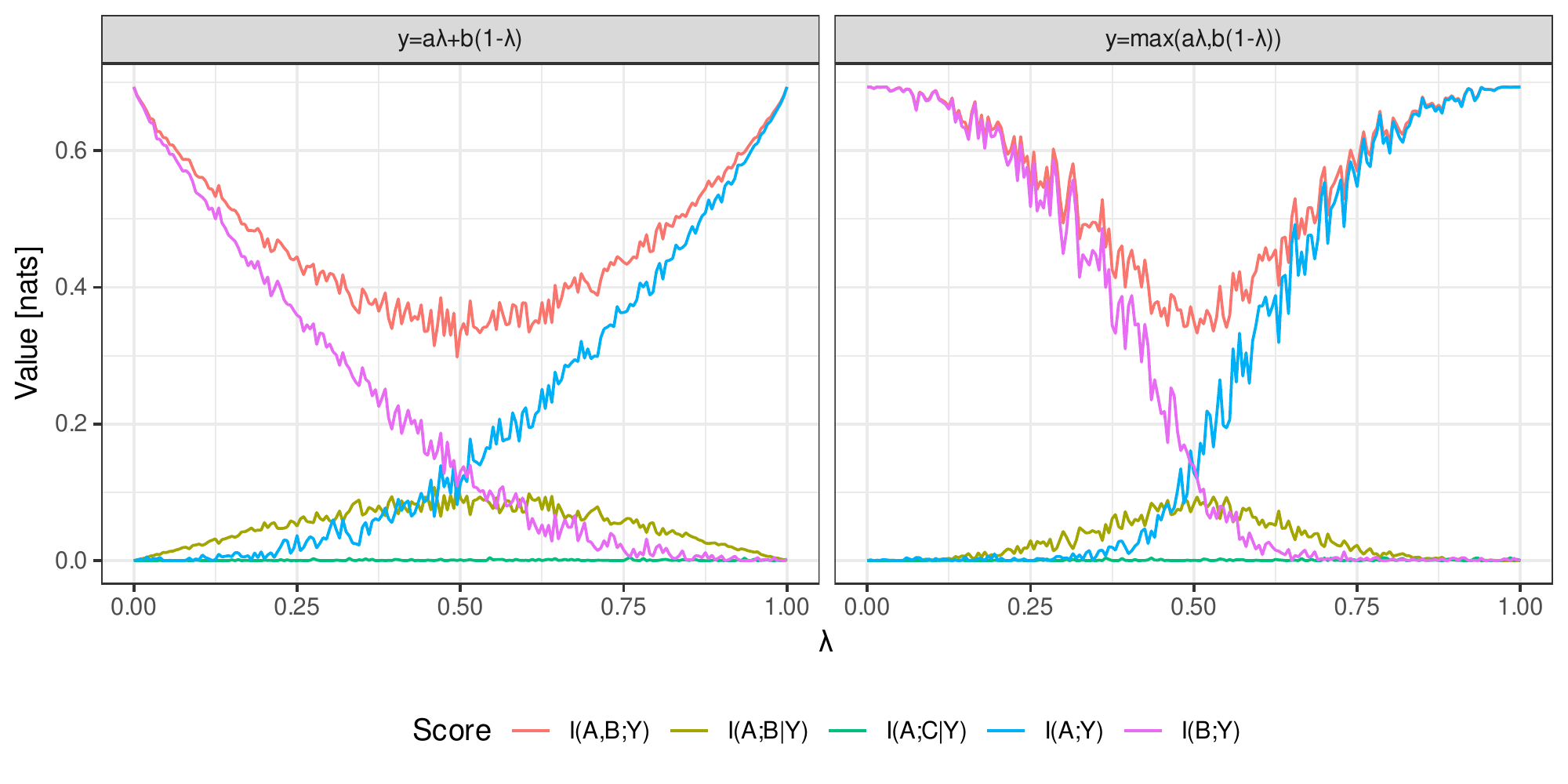}
 \caption{\label{fig:jmi} Values of certain information scores for a Kendall-transformed system of three features $a$, $b$ and $y$ engaged in a simple interaction, $y=a\lambda+b(1-\lambda)$ (left) or $y=\max\{a\lambda,b(1-\lambda)\}$ (right), for a range of realisations with a different $\lambda$ parameter.}
\end{figure}
The important gain of Kendall transformation is that transformed features can be used in more complex considerations that just bivariate ones.
In particular, we can calculate joint, conditional or multivariate mutual information and use it to investigate relationships between features.

For example of such an analysis, let us consider an information system composed of three independent, random features $a,b,c\sim U(0,1)$ and a decision which is either a linear $y=a\lambda+b(1-\lambda)$ or an example non-linear $y=\max(a\lambda,b(1-\lambda))$ function of $a$ and $b$.
It is worth noticing that although linear relation seems pretty basic, Kendall transformation makes it invariant to monotonic transformations of $a$, $b$ and $y$, hence this case covers more complex functions, for instance $y=\sinh^\mu(a)\cdot{(b-b_0)}^\nu$.
Let us also denote the Kendall-transformed features with upper-case letters, i.e., $A=\K(a)$ and so on.
Figure~\ref{fig:jmi} contains the maximum likelihood estimates of certain information scores in such system, for $n=200$ and a selection of realisations for a range of $\lambda$ values.

We can see that for most realisations the joint mutual information $I(A,B;Y)$ is larger than either marginal mutual information, signifying that considering both features allows for a better prediction of $Y$.
Their interaction can be directly measured with conditional mutual information $I(A;B|Y)$; we can also compare it to a baseline of $I(A;C|Y)$, which is asymptotically zero as $C$ is irrelevant.
This score substantially dominates baseline for almost full range of $\lambda$ in the linear case, and at least half of it in the much harder non-linear case, confirming the previous conclusion of the presence of interaction.
Moreover, it reaches maximum for $\lambda=1/2$, i.e., balanced impacts of $a$ and $b$, and decreases as either of them dominates, which is a reasonable outcome.

In a yet another view, we can analyse the three-feature mutual information $I(A;B;Y)$.
Because of the independence of $A$ and $B$, it is approximately equal to $-I(A;B|Y)$, hence has a negative minimum for $\lambda=0$, which also signifies synergy between the three features.
Similarly, the marginal MIs $I(A;Y)$ and $I(B;Y)$ behave as a reasonable measure of relative impact, akin weights in linear regression, yet generalisable over many non-linear relationships.

The other important aspect of using Kendall transformation for multivariate analysis is how well we can estimate the joint distribution, which generally depends on how many states are there, and how many observations of each can we count.
The number of states increases exponentially with the dimension of the analysis, yet the base of said exponent is usually $2$ when using Kendall transformation and $5-10$ in case of discretisation.
Also, due to the nature of the Kendall transformation, the effective number of observations is squared and marginal distributions are well balanced, reducing the noise in counts and the risk of spuriously unobserved states.

Kendall-transformed information systems can also be used as an input for sophisticated machine learning methods.
Some approaches can operate in a straightforward way, like many feature selection methods which may be used for explanatory analysis, others may require appropriate adaptations.
The particular caveat is that the transformation generates artificial correlation structure between pairs, hence methods which rely on them being independent may become biased.

\subsection{\label{sec:it} Inverse transformation}
A Kendall-transformed feature can be transformed back using the following method.
First, for each object $i$ we extract values of pairs $(i,\cdot)$, and score each $\U$ with $1$ and each $\D$ with $-1$.
Then, we order objects according to the decreasing score, which recovers the ranking.
Such algorithm is equivalent to the Copeland's election method~\cite{Copeland1951}, and it is trivial to see that it always correctly reverses a valid result of a Kendall transformation.

Kendall transformation is non-surjective, henceforth, when operating on transformed data, we can stumble upon invalid sequences of $\U$, $\D$ and $\E$.
For instance, let's consider a feature $(\U,\D,\D,\U,\U,\D)$ --- it corresponds to a cycle of three observations, and cannot be generated by transforming any ranking.
While there is no universal solution covering such cases, Copeland's method can be easily adapted and used as one of possible heuristic solutions; in this particular case, it maps these features into all observations with the same rank, which intuitively seems as a reasonable solution.

Furthermore, sometimes we can also expect missing values or fuzzy predictions, i.e., probabilities or weights of each state.
In this cases, we can reach out deeper into the social choice theory for a more sophisticated approach, like the Schulze's~\cite{Schulze2011} or Tideman's~\cite{Tideman1987} method.
They will produce a consensus graph of ordering relations, from which we can recover complex tie structures and disconnected subsets of observations.

\subsection{\label{sec:mg} Merging of transformed data}

One of the common problems in molecular biology is the lack of consistent calibration of certain experimental procedures, especially the high-throughput ones.
From a mathematical point of view, this phenomenon can be well modelled as an assumption that results for each batch of samples are filtered by an arbitrary monotonic function, specific to a given batch.
This way, explanatory outcomes from different batches are mostly consistent; yet, raw datasets cannot be naively combined, for instance to look for more subtle aspects in meta-analysis.
It is because the batch effect can be very substantial, even to the point of overshadowing actual interactions, heavily biasing the results.

Kendall transformation can be used as another approach to tackle this problem.
The independently transformed data sets from different batches do not retain the strictly increasing dis-calibration effects, hence they can be safely combined without the risk of introducing bias.
The merged data will not represent relations in cross-dataset pairs, but we can treat these rows as missing and still conduct actions like feature selection or model training.
Only for the prediction, the sets have to be separated to perform inverse transformation --- the cross-set relations will be left unknown, but so they were in the first place.

\section{Example}
For an example analysis using the Kendall transformation, let us consider the morphine withdrawal data set from~\cite{Hamed2018}.
It collects concentrations of 15 neurotransmitters in 6 brain structures, measured in four groups of rats: subject to either a morphine or saline treatment, as well as measured directly after treatment or after a 14 day withdrawal period and re-exposure to the administration context.
Additionally, ultrasonic vocalisation (USV) intensity of rats was also quantified as a number of episodes during a 20 minute recording session.
In contrast to the original, in the data used here one record has been rejected due to missing data.

Overall, the set is composed of 37 objects and 90 continuous features, distributions of which are predominantly not normal, as well as three separate decision features, continuous USV episode count, and two categorical: morphine treatment and withdrawal period.
Such structure is typical to many biomedical studies.

\subsection{Feature ranking}
\begin{table}[ht]
\centering
\begin{tabular}{rccc}
 & USV & Morphine & Withdrawal \\ 
  \toprule
Kendall transformation & 1.00 & 1.00 & 1.00 \\ 
Three equal-width bins & 0.71 & 0.50 & 0.56 \\ 
Five equal-width bins & 0.83 & 0.50 & 0.62 \\ 
Three equal-frequency bins & 0.44 & 1.00 & 0.86 \\ 
Five equal-frequency bins & 0.40 & 0.17 & 0.64 \\ 
\midrule
Random Forest importance & 0.80 & 1.00 & 0.86 \\ 
\end{tabular}
\caption{\label{tab:ag} Agreement of mutual information feature rankings obtained using different data transformations with significant results of a standard non-parametric statistical analysis applied to the morphine withdrawal data set, given as a maximal value of the Jaccard index.
Random Forest importance ranking added for comparison.}
\end{table}

In the original paper, a standard, bivariate non-parametric statistical analysis was used to identify compounds significantly connected with each of the decision features.
Namely, categorical decisions were analysed with Mann-Whitney-Wilcoxon $U$ test~\cite{Mann1947}, while continuous one with a Spearman $\rho$ test~\cite{VanDeWiel2001}.
Such analysis, combined with a multiple comparisons correction, yields 5 significant features for the episode count problem, 1 for morphine and 13 for withdrawal.
Let us compare such outcome with the mutual information rankings of features obtained on either Kendall-transformed or binned data, as well as with the Random Forest importance~\cite{Breiman2001} applied directly (using the randomForest R package~\cite{Liaw2002}, in its default set-up).

Table~\ref{tab:ag} collects the results of the said experiment; agreement is quantified by the maximal value of Jaccard index \cite{Jaccard1901} over all possible cut-offs in the respective ranking.
We see a perfect agreement in case of Kendall transformation, which is unsurprising given the aforementioned equivalence relations, as well as the fact that Spearman and Kendall correlation coefficients are usually highly correlated.
The rankings on discretised data are substantially influenced by the binning method; even though all predictors are the same, a different method is optimal for each decision.
Perfect agreement with baseline is only achieved once, and the agreement is pretty poor on average.
The much more elaborated approach, RF importance, achieves a relatively high average agreement of 0.89, given that it also considers multivariate relations.
Overall, the results support the notion that simple discretisation is susceptible to exaggerating spurious interactions inherent to small sample data.
Kendall transformation not only helps to avoid this phenomenon, but also requires no hyper-parameters, which have a critical impact on binning.

\subsection{Prediction}
\begin{table}[ht]
\centering

\begin{tabular}{rccc}
& USV [Spearman cc.] & Morphine [AUROC] & Withdrawal [AUROC] \\
  \toprule
Transformed data & 60.1\% [46.1\%-70.1\%] & 82.8\% [71.4\%-89.3\%] & 100.0\% [98.1\%-100.0\%] \\
  Original data & 62.2\% [49.2\%-69.7\%] & 87.1\% [76.5\%-96.8\%] & 100.0\% [97.4\%-100.0\%] \\ 
\end{tabular}
\caption{\label{tab:cls} Test-set accuracy of a Random Forest classifier trained on the morphine withdrawal data set, Kendall-transformed and original.
Values provided as median [IQR] over 100 simulation realisations.}
\end{table}

In order to investigate Kendall transformation in the context of machine learning, let us now compare the accuracy of the Random Forest method applied to the morphine data directly and after Kendall transformation.
Accuracy is estimated using a bootstrap train-test split, hence with roughly 63\% of observations used to build a model predicting the remaining data -- in both cases the split is done before transformation.
For the modelling after Kendall transformation, predictions for pairs are translated to ranking of original objects using the Copeland's method proposed in Section~\ref{sec:it}, considering the fraction of votes cast for the $\U$ state.
Similarly, for the baseline, the vote fraction is used as an outcome for classification problems (Morphine and Withdrawal), while the regression mode is used for the USV count prediction.
Finally, the accuracy is quantified with Spearman correlation, for regression, or with AUROC, for classification.
As in previous example, the randomForest~\cite{Liaw2002} implementation is used.

The result of this analysis, repeated over 100 random replications, is reported on Table~\ref{tab:cls}.
We see that the application of Kendall transformation, despite its lossy nature, does not substantially hinder the accuracy of the prediction; the loss is statistically significant only in case of the Morphine problem, yet it is about a quarter of the interquartile range.
Thus, we can conclude that Kendall-transformed data is a viable input for a machine learning method.

\subsection{Data integration}
For the sake of this example, let us simulate the acquisition of data from two incoherent measurements.
To this end, the morphine set is randomly split in half, and the values of features in one part is tripled.
Next, we either naively merge halves back and apply Kendall transformation on the fused set, or the other way round, merge independently transformed parts.
Finally, we compare the agreement of mutual information-based feature rankings obtained after each processing with this calculated on the golden standard, Kendall-transformed original, unperturbed data.

\begin{table}[ht]
\centering
\begin{tabular}{rccc}
 & USV & Morphine & Withdrawal \\
  \toprule
Transformed merge & 95.1\% [93.5\%-96.0\%] & 96.0\% [93.8\%-96.9\%] & 98.3\% [97.3\%-98.7\%] \\
Naive merge & 26.6\% [24.5\%-39.4\%] & 41.8\% [-23.8\%-47.5\%] & 71.7\% [21.6\%-91.2\%] \\
\end{tabular}
 \caption{\label{tab:agm} Robustness of Kendall transformation-based feature ranking to a simulated loss of calibration between two measurement sessions, quantified by Spearman correlation with the ranking on unperturbed data, using the morphine withdrawal example.
 Values provided as median [IQR] over 100 simulation realisations.}
\end{table}

The results, quantified by Spearman correlation coefficient and averaged over 100 realisations, are presented as Table~\ref{tab:agm}.
We see that the simulated measurement artefact has a substantial, negative impact on the accuracy and stability of the ranking in the naive scenario, while the merge of Kendall-transformed parts reliably achieves over 90\% agreement.
This is expected due to a fact that the applied disruption is a strictly increasing function, which Kendall transformation is invariant to.
The agreement is not perfect, however, because the set merged after transformation lacks almost 50\% of information, namely all the pairs representing intra-half relationships --- in light of this fact, the observed loss is rather limited.

\section{Conclusions}
Kendall transformation is a novel way to represent ordinal data in a categorical form, as well as to apply discrete methods and approaches on such data.
While standard discretisation procedures sacrifice precision, Kendall transformation precisely preserves ranking, sacrificing original distribution instead.
This approach is common to non-parametric statistical methods, though, and it is proved effective in many use-cases, in particular in small sample size conditions.
In fact, I show that Kendall transformation is tightly connected with certain popular methods of this class on a theoretical level.

Moreover, Kendall transformation is reversible into ranking, has no parameters and imposes no restrictions on the input, as well as offers consistent behaviour regardless of its characteristics.
The method is also very versatile, as its output can be used both to calculate some simple coefficient and be a part of an elaborate algorithm.

\bibliography{text}

\section{Availability}
The implementation of Kendall transform (and its inverse) for R is available in CRAN package praznik\cite{Kursa2021}, starting from version 8.0.
The code used to execute the computational experiments presented in the paper can be found on GitHub, \url{https://github.com/mbq-suppl/kendall-transformation}.

\end{document}